\documentclass{article}

\PassOptionsToPackage{numbers, compress}{natbib}
 \usepackage[preprint]{neurips_2026}

\usepackage[utf8]{inputenc} 
\usepackage[T1]{fontenc}    
\usepackage{hyperref}       
\usepackage{url}            
\usepackage{booktabs}       
\usepackage{amsmath}        
\usepackage{amsfonts}       
\usepackage{nicefrac}       
\usepackage{microtype}      
\usepackage{xcolor}         
\usepackage{graphicx}       
\usepackage{multirow}
\usepackage{float}          
\graphicspath{{figures/}}
\newcommand{\MISSING}[1]{\textcolor{red}{TBD}}

\title{The Position Curse: LLMs Struggle to Locate the Last Few Items in a List}

\author{
  Zhanqi Zhang\thanks{Co-first authors} \\
  UC San Diego \vspace{-2pt} \\
  \texttt{zhz091@ucsd.edu} \\
  \And
  Hua-Dong Xiong\footnotemark[1] \\
  Georgia Tech \vspace{-2pt} \\
  \texttt{hdx@gatech.edu} \\
  \And 
  Robert C. Wilson \\
  Georgia Tech \vspace{-2pt} \\
  \texttt{rwilson337@gatech.edu} \\
  \And
  Mikio Aoi \\
  UC San Diego \vspace{-2pt} \\
  \texttt{maoi@ucsd.edu} \\
  \And
  Marcelo G. Mattar \\
  New York University \vspace{-2pt} \\
  \texttt{marcelo.mattar@nyu.edu} \\
  \And
  Li Ji-An\thanks{Senior author.} \\
  New York University \vspace{-2pt} \\
  \texttt{jl9246@nyu.edu} \\
}
 
\begin{document}

\maketitle

\begin{abstract}  
Modern large language models (LLMs) can find a needle in a haystack (locating a single relevant fact buried among hundreds of thousands of irrelevant tokens) with near-saturated accuracy, yet fail to retrieve the last few items in a short list. We call this failure the \emph{Position Curse}. For instance, even in a two-line code snippet, Claude Opus 4.6 misidentifies the second-to-last line most of the time. To characterize this failure, we evaluated two complementary queries: given a position in a sequence (of letters or words), retrieve the corresponding item; and given an item, return its position. Each position is specified as a forward or backward offset from an anchor, either an endpoint of the list (its start or end) or another item in the list. Across both open-source and frontier closed-source models, backward retrieval substantially lags forward retrieval. To test whether this capability can be rescued by post-training, we constructed \textsc{PosBench}, a position-focused training dataset. LoRA fine-tuning improves both forward and backward retrieval and generalizes to a held-out code-understanding benchmark (\textsc{PyIndex}), yet absolute performance remains far from saturated. As LLM coding agents increasingly operate over large codebases where precise indexing becomes essential for code understanding and editing, position-based retrieval emerges as a key capability for future pretraining objectives and model design.
\end{abstract}
   
\vspace{-0.12in}
\begin{figure}[H]
    \centering
    \includegraphics[width=0.75\linewidth]{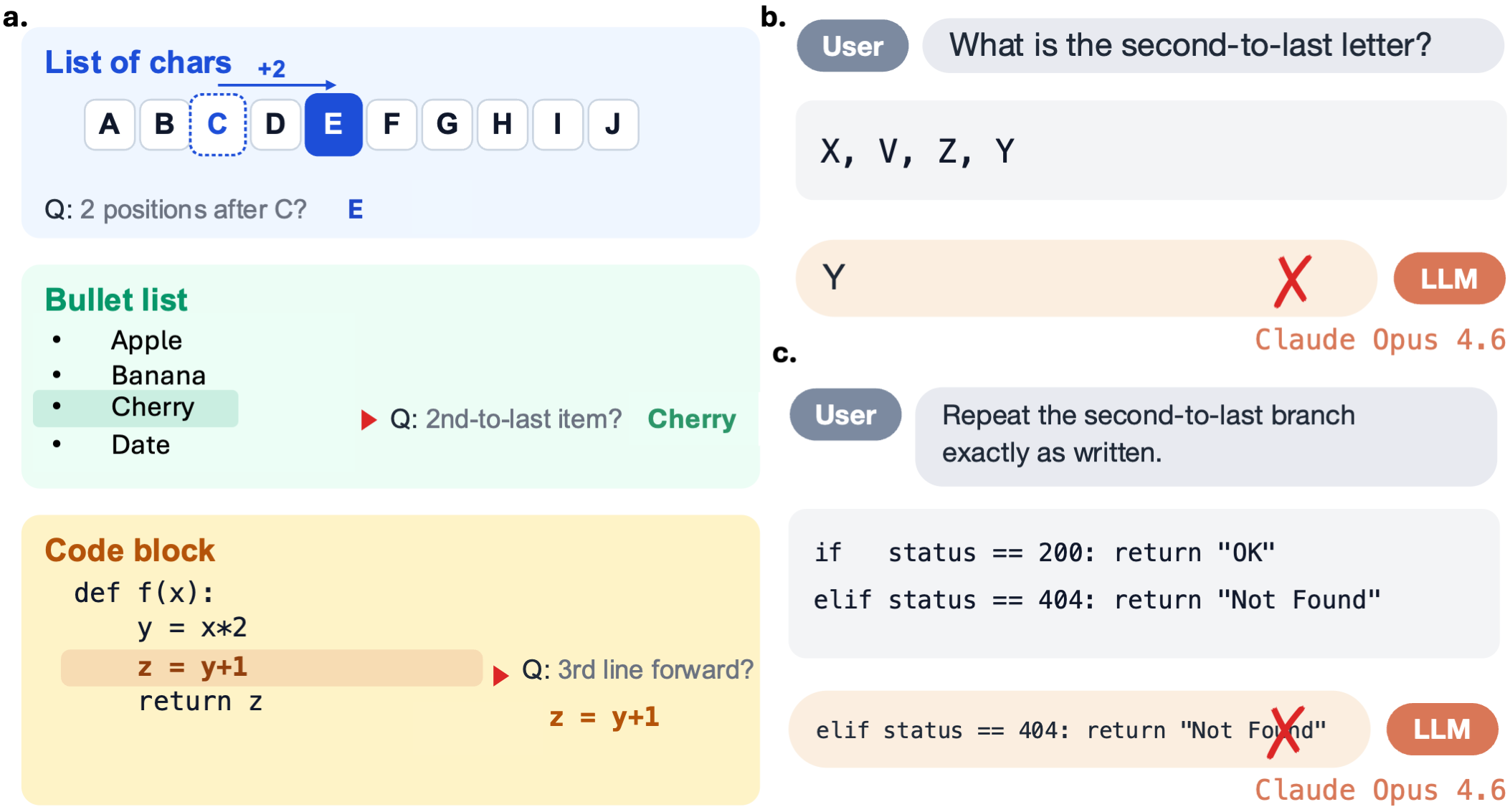}
    \caption{\textbf{Position-based retrieval.}
    (a) Index-based query formats over short sequences: letter list, bullet list, and code block.
    (b,c) Example failures by Claude Opus 4.6; 
    full prompts in Appendix~\ref{app:fig1-prompts}.}
    \label{fig:fig1}
\end{figure}
\vspace{-0.12in}

\section{Introduction}\label{sec:intro}

Modern LLMs are increasingly deployed as coding agents, such as Claude Opus 4.6 powering Claude Code, where they must track multiple code snippets across long conversations while tracing bugs, planning refactors, and extending code across many files and interaction turns. This usage pattern is becoming increasingly common as frontier models support context windows approaching one million tokens~\citep{anthropic2026opus47,openai2026gpt55}. As increasingly large codebases and documents fit into context, the bottleneck shifts from encoding information in the context window to reliably retrieving precise, multi-level information from it. Existing long-context evaluations, including Needle in a Haystack~\citep{kamradt2023needlehaystack}, RULER~\citep{hsieh2024rulerwhatsrealcontext}, LongBench~\citep{bai2024longbench}, $\infty$Bench~\citep{zhang2024infinitebench}, and Graphwalks~\citep{openai2025graphwalks}, primarily focus on \emph{content-based} retrieval, aiming to match a query to its semantically relevant context. Frontier models now approach saturation on many of these content-based retrieval tasks.

A different capability, \emph{position-based} retrieval, also known in historical memory research as \emph{location-based addressing}, is largely ignored by current long-context benchmarks, with MRCR as a rare exception~\citep{openai2025mrcr}. Classical computer architecture~\citep{patterson2017computer} and neural memory models, such as Neural Turing Machines~\citep{graves2014neuralturingmachines}, treat position-based retrieval as a fundamental atomic operation. The Transformer architecture likewise incorporates positional information through positional encodings~\citep{vaswani2017attention}, highlighting the importance of supporting position-based access. Position-based retrieval also appears throughout everyday and coding workflows: summarizing the first or last few paragraphs of a document, inspecting the most recent entries in a log file, or understanding indexing and slicing expressions in a code block.

We investigated whether open-source and proprietary frontier LLMs can perform position-based retrieval. Given a short list of items, we asked the model to retrieve the item at the queried $n$th position from the start, from the end, or relative to a given item; and to report the position at which a queried item appears (Fig.~\ref{fig:fig1}a). Surprisingly, nearly all evaluated models struggle on these queries even at trivial sequence lengths, with the performance of backward retrieval consistently lagging forward retrieval. We call this systematic weakness the \emph{Position Curse}. Two representative failures from Claude Opus 4.6, which drives the popular coding agent Claude Code, are shown in Fig.~\ref{fig:fig1}: given a four-letter list, the model misidentifies the second-to-last letter 27\% of the time (Fig.~\ref{fig:fig1}b); given a two-line code snippet, it misidentifies the second-to-last, i.e., the first, line (Fig.~\ref{fig:fig1}c).

\begin{figure}[htbp]
    \centering
    \includegraphics[width=0.85\linewidth]{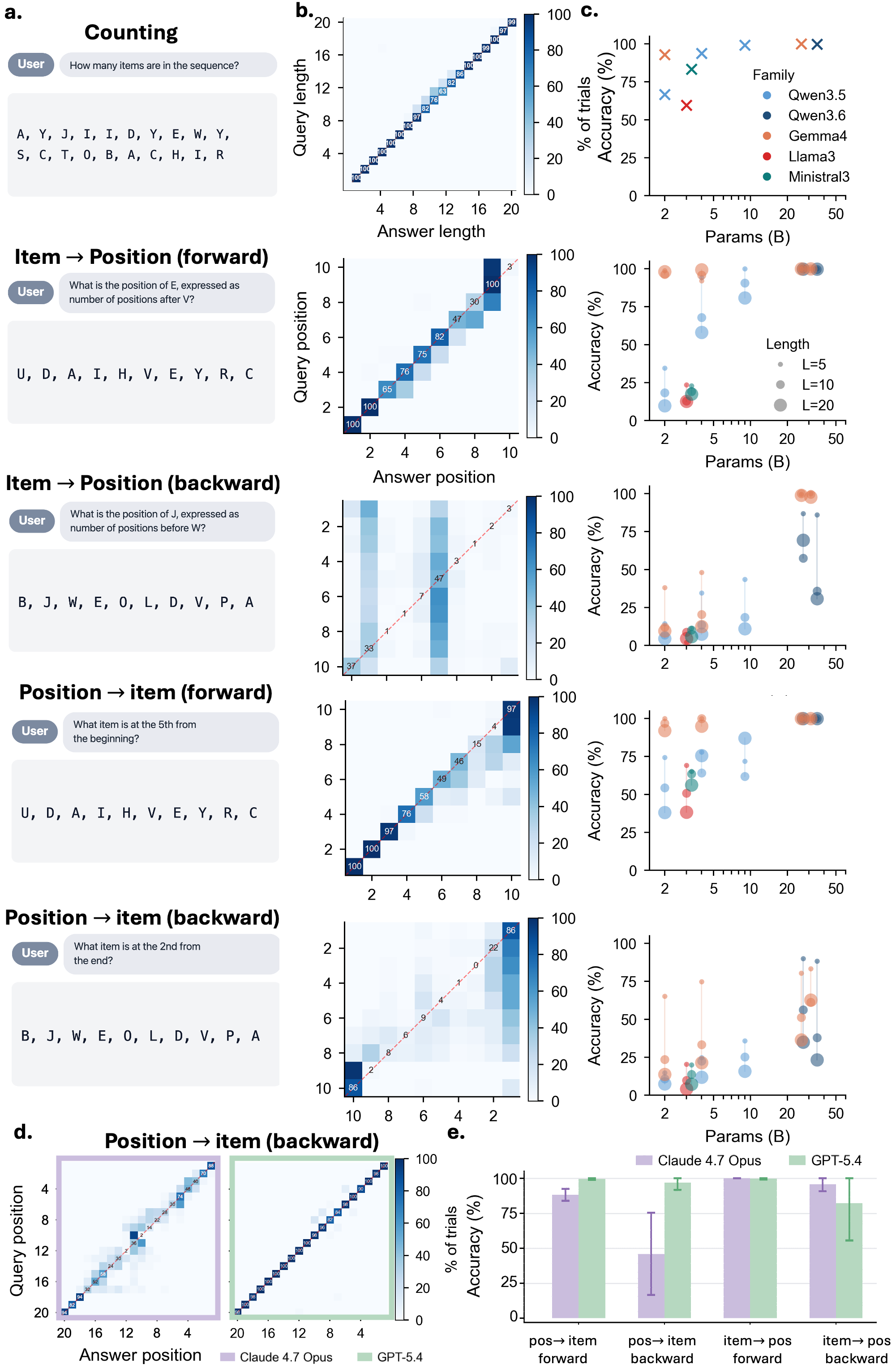}
    \caption{\textbf{Position-based retrieval across tasks and models.}
    \textbf{(a)} Example prompts for the counting control and the four retrieval variants (position$\rightarrow$item and item$\rightarrow$position, each queried forward and backward).
    \textbf{(b)} Qwen3.5-4B confusion heatmaps (letters, $L{=}10$). Cell color is the row-normalized percentage of trials in each answer bin (length for counting; position for retrieval); the red dashed diagonal marks the correct answer. For backward queries, axes are ordered by absolute position, so queried values decrease top-to-bottom and answered values decrease left-to-right.
    \textbf{(c)} Accuracy across model scales. Each point is the mean exact-match accuracy for one (model, task, sequence-length) condition, averaged over valid query positions and trials. Color denotes model family; marker size denotes sequence length.
    \textbf{(d)} Position$\rightarrow$item confusion for Claude 4.7 Opus and GPT-5.4 (letters, $L{=}20$, endpoint anchor); same heatmap convention as (b).
    \textbf{(e)} Mean position-based retrieval accuracy for Claude 4.7 Opus and GPT-5.4. Bars show average accuracy across the 20 queried positions for each (task, model); error bars are the SD across those per-position accuracies.}
        
    \label{fig:fig2}
\end{figure}

These failures might seem to relate to LLMs' known difficulty with counting (e.g., counting letters in a word)~\citep{fu2024largelanguagemodelsllms,zhang2024countingabilitylargelanguage,cosma2025strawberry}. Yet, as we showed in Sec.~\ref{sec:results_count} and Fig.~\ref{fig:fig2} (row 1), empirical observations suggest otherwise. Counting performance on these lengths is far higher than retrieval performance on the same lists. The \emph{Position Curse} is therefore a distinct phenomenon in which the model can count, yet fails to retrieve the (relative) position of the queried item.

Our main contributions are as follows. First, we identified and systematically characterized the \emph{Position Curse} across four axes: two query directions (position$\rightarrow$item: given a position, retrieve the item there, and item$\rightarrow$position: given an item, return its position), two anchor types (relative to the start/end of the list, and relative to a given item), two indexing directions (forward and backward), and two types of items (single-token letters and multi-token words). Across all axes, the performance of backward retrieval lags forward retrieval. Second, to test whether post-training can rescue position-based retrieval, we constructed \textsc{PosBench}, a position-focused training dataset, and showed that low-rank adaptation (LoRA)~\citep{hu2022lora} reliably yields improvements on the in-distribution tasks. Third, we introduced \textsc{PyIndex}, a synthetic benchmark designed to evaluate position-based retrieval in code understanding, and found the fine-tuned capability generalizes reliably to the held-out \textsc{PyIndex} code-understanding benchmark, though absolute performance remains far from saturated. Together, these results show that position-based retrieval is a distinct capability gap inherited from pretraining and an important target for training objectives, data curation, and model design.

\section{Related Work}

In addition to the works discussed in Sec.\ref{sec:intro}, two prior lines of work touch on positional effects in retrieval. \emph{Lost in the Middle}~\citep{liu2023lostmiddlelanguagemodels} measures how the position of the target in a long context affects retrieval accuracy, reporting a characteristic U-shape positional effect in retrieval. However, its retrieval remains content-based rather than position-based retrieval studied here. OpenAI's MRCR~\citep{openai2025mrcr} is a rare long-context benchmark involving a positional component: 2, 4, or 8 identical user requests are embedded in a long synthetic dialog, and the model is asked to retrieve ``the $i$-th'' instance. However, MRCR  does not systematically vary direction (forward vs.\ backward), anchor (start/end of the list vs.\ relative to an item), query type (position$\rightarrow$item vs.\ item$\rightarrow$position), or item types (letters/words), as we do here.

\section{Results}\label{sec:results}

We begin by introducing the tasks studied in this section with a length-counting control task, and a family of position-based retrieval tasks.

All queries were presented to the model in a three-shot format, where each test prompt is preceded by three in-context demonstrations of the same task instantiated on independently sampled sequences. We adopted few-shot rather than zero-shot prompting to disambiguate the task itself, so that any failure reflects the model's ability on the task at hand, whether length counting or position-based retrieval, rather than ambiguity about the query/answer format (e.g., 0- vs. 1-based indexing, or how to phrase the answer).

\subsection{Length counting performance is near-ceiling}\label{sec:results_count}

A natural concern is that failures on position-based retrieval may reflect a more basic deficit, namely that models cannot even count the items in a sequence. To rule this out, we use length counting as a minimal control task. Given a sequence, the model was asked ``How many items are in the sequence?'' (Fig.~\ref{fig:fig2}a, row 1). Unlike the retrieval tasks introduced later, counting here does not require retrieving items at specific positions, but still requires tracking sequence length,  isolating general counting from positional retrieval.

Fig.~\ref{fig:fig2}b, row 1 visualizes counting on Qwen3.5-4B as a confusion heatmap. The x-axis is the length returned by the model, the y-axis is the queried sequence length, cell intensity is the percentage of trials, and the red dashed diagonal marks the correct length. The mass is concentrated tightly on the diagonal, suggesting that the model returns the queried length almost without error across the tested range.

As shown in Fig.~\ref{fig:fig2}c, row 1, this near-ceiling behavior is consistent across model sizes. Qwen3.5 achieves 78\% accuracy at 2B, 96\% at 4B, and 100\% at 9B, while larger models remain at ceiling. Across sequence lengths up to $L=20$, accuracy remains between 94\% and 100\%, indicating that simple length tracking is effectively saturated.

\subsection{Setup: tasks, prompting, and performance evaluation}\label{sec:results_setup}

We formalize position-based retrieval as a deterministic operation over an ordered sequence. This formulation separates the structure of the query from the surface form of the prompt, allowing the same underlying operation to be tested across four task axes, as well as across model families. The first axis is \emph{query direction}. Position$\rightarrow$item means given a position retrieve the item there, whereas item$\rightarrow$position means given an item return its position. The second axis is \emph{anchor type}. The anchor may be either an endpoint of the list or another item in the list. The third axis is \emph{indexing direction}. Indexing may proceed either forward or backward from the anchor. The fourth axis is \emph{item type}. Items may be either single-token letters or multi-token words. The main text defines the task family needed to interpret the experiments, while Appendix~\ref{app:data} provides the full model coverage, dataset construction, training methods, and performance evaluation.

Let $S=[s_1,\ldots,s_L]$ be an ordered sequence. A query specifies an offset $n\ge 1$, an anchor (either a sequence endpoint, denoted $\mathsf{End}$, or an item at position $r$, denoted $\mathsf{Rel}$), and a direction (superscript $+$ for forward, $-$ for backward). The four index operators are
\[
\mathsf{End}^{+}\!: S[n],\qquad \mathsf{End}^{-}\!: S[-n],\qquad \mathsf{Rel}^{+}\!: S[r{+}n],\qquad \mathsf{Rel}^{-}\!: S[r{-}n],
\]
with $S[-n]\equiv s_{L-n+1}$ and offsets sampled so that the index lies in $[1,L]$. Position$\rightarrow$item queries return the indexed item; item$\rightarrow$position queries invert the same operator and return the offset $n$ given a target item. We instantiated this family across letters and words, endpoint and relative anchors, both directions, and sequence lengths $L\in\{5,10,20\}$, shown in Fig.~\ref{fig:fig2}a (row 2--4) and Appendix Fig.~\ref{fig:figA2}a. 

To illustrate the operator family, here are example position$\rightarrow$item prompts for each operator. $\mathsf{End}^{+}$ asks ``What item is at the 3rd position from the beginning?'', $\mathsf{End}^{-}$ asks ``What item is at the 2nd position from the end?'', $\mathsf{Rel}^{+}$ asks ``What item is two positions after $X$?'', and $\mathsf{Rel}^{-}$ asks ``What item is one position before $X$?''. The answer space is determined by the query direction. For position$\rightarrow$item queries, the answer is one of the items in $S$; for item$\rightarrow$position queries, the answer is the integer offset $n$. Model responses are parsed into this answer space and scored by exact match.

\subsection{Forward-backward asymmetry persists across scales and families}\label{sec:results_pos}
  
Having ruled out a basic counting deficit, we now turn to the four position-based retrieval tasks. We first examine the per-task behavior of a single model, Qwen3.5-4B (Fig.~\ref{fig:fig2}b), and ask whether the same patterns hold across model scales and families (Fig.~\ref{fig:fig2}c).

Fig.~\ref{fig:fig2}b reports per-task confusion heatmaps for Qwen3.5-4B on the four position-based retrieval tasks. The x-axis is the position answered by the model; the y-axis is the queried position. The two forward retrieval tasks (rows 2 and 4) are roughly concentrated along the diagonal but with visible off-diagonal mass, indicating that the model tracks positions reasonably but not error-free. The two backward retrieval tasks (rows 3 and 5) instead collapse into vertical stripes, where the model lands on a small set of canonical answers regardless of the queried backward position.

Fig.~\ref{fig:fig2}c shows that this asymmetry is not idiosyncratic to Qwen3.5-4B. It reports accuracy across the same four retrieval tasks for Qwen3.5 (2B, 4B, 9B), Qwen3.6 (35B-A3B), and Gemma4 (E2B and 26B-A4B), Llama32-3B and Ministral3-3B as a function of model size and sequence length $L \in \{5,10,20\}$. The forward--backward asymmetry persists across all of these families.

Forward endpoint retrieval ($\mathsf{End}^{+}$) is the easiest retrieval setting. On Qwen3.5, $L=20$ accuracy rises from 22.9\% at 2B to 83.9\% at 9B, and larger Gemma and GPT models approach ceiling performance in this setting. In contrast, backward endpoint retrieval ($\mathsf{End}^{-}$) remains fragile. For position$\rightarrow$item queries of the form ``$n$th from the end'', Qwen3.5 improves only from 5.8\% at 2B to 16.1\% at 9B.

Relative-anchored item$\rightarrow$position ($\mathsf{Rel}^{\pm}$, Fig.~\ref{fig:fig2}c, rows 2--3) is also difficult for the Qwen3.5 dense models. At $L=20$, performance reaches only 15.2\% at 9B and remains far below the corresponding $\mathsf{End}^{+}$ setting. Although other model families perform better, the Qwen trend suggests that parameter scale alone does not automatically repair relative indexing.

Together, these results identify $\mathsf{End}^{-}$ and $\mathsf{Rel}^{\pm}$ as the central failure modes. Larger models reach high accuracy on $\mathsf{End}^{+}$, but the same models remain unreliable when the index must be transformed through backward indexing or through a relative anchor.

\subsection{Reasoning alone does not resolve the Position Curse}\label{sec:results_thinking}

\begin{figure}[htbp]
    \centering
    \includegraphics[width=\linewidth]{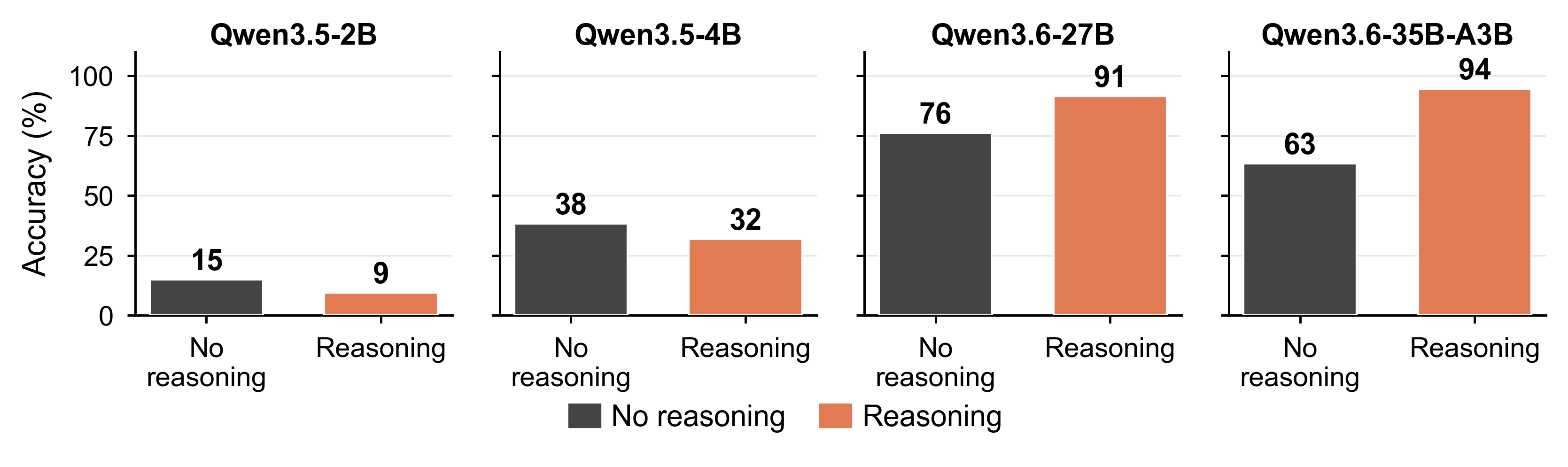}
    \caption{\textbf{Mean accuracy under reasoning vs.\ no-reasoning.} Bars show mean accuracy (\%) on letter-sequence, endpoint-anchored position retrieval, averaged across the four position-retrieval tasks and across sequence lengths $L$. Model panels are ordered from left to right by increasing capacity: Qwen3.5-2B, Qwen3.5-4B, Qwen3.6-27B, and Qwen3.6-35B-A3B. Within each panel, gray bars show the no-reasoning condition and orange bars show the reasoning condition. See Appendix Fig.~\ref{fig:figA1} for detailed per-position results.}
    \label{fig:fig3}
\end{figure}

The results so far were obtained under the \emph{no-reasoning} condition. The model emits its answer directly, without producing any intermediate reasoning tokens. We then ask whether allowing the model to think before answering closes the forward--backward gap.

Four position-retrieval tasks were evaluated at $L{=}20$ (letters, endpoint anchor) for five Qwen variants under both \emph{no-reasoning} and \emph{reasoning} conditions. In the reasoning condition, the model was allowed up to 256 intermediate reasoning tokens before producing its final answer. 

Reasoning did not uniformly improve positional retrieval. Overall, the larger Qwen3.6 models benefited substantially more from reasoning than the smaller Qwen3.5 models (Fig.~\ref{fig:fig3}). For the smaller Qwen3.5-2B and 4B models, reasoning left performance largely unchanged or slightly degraded retrieval accuracy across most tasks and query positions (Appendix Fig.~\ref{fig:figA1}, cols.~1--2).
 
The Qwen3.6 models, in contrast, showed large gains, concentrated almost entirely in the backward conditions where the no-reasoning condition exhibited broad mid-sequence failures (Appendix Fig.~\ref{fig:figA1}, cols.~3--4). For example, Qwen3.6-35B-A3B improved from $63\%$ to $94\%$ accuracy under reasoning. In the position$\rightarrow$item backward setting, reasoning lifted the mid-sequence accuracy that had dipped in the no-reasoning curves (Appendix Fig.~\ref{fig:figA1}, row 2, cols.~3--4), while in the item$\rightarrow$position backward setting, it largely eliminated the severe middle-position collapse observed in the no-reasoning condition (Appendix Fig.~\ref{fig:figA1}, row 4, cols.~3--4). However, even with reasoning, the Qwen3.6 models continued to make frequent errors on backward retrieval.

Together, these results suggest that reasoning helps larger models but remains insufficient to solve position-based retrieval. Moreover, requiring large amounts of intermediate reasoning tokens on a capability this basic is itself inefficient.

\subsection{LoRA Fine-tuning reliably improves position-based retrieval}\label{sec:results_recipe}

\begin{figure}[t]
  \centering
  \includegraphics[width=\linewidth]{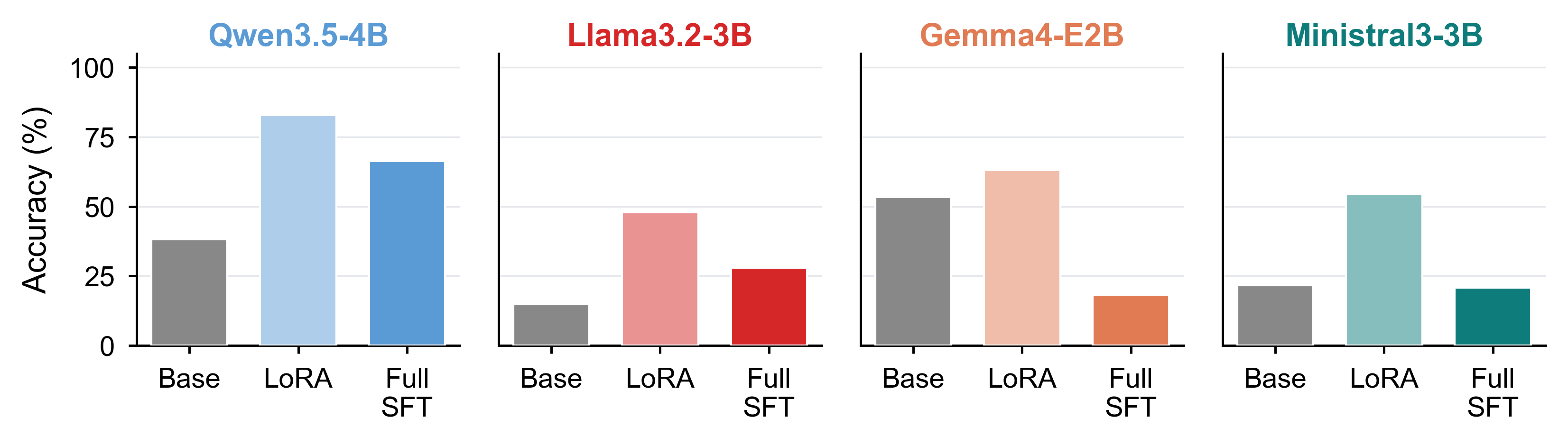}
  \caption{\textbf{Position-based retrieval accuracy with fine-tuning.}
  Accuracy across four open-source model families (Qwen3.5-4B, Llama 3.2-3B, Gemma4 E2B, Ministral 3-3B) under two fine-tuning methods (LoRA, full-parameter supervised fine-tuning SFT). Bars show mean exact-match accuracy on letter sequences with endpoint anchoring at $L{=}20$, averaged across the four position-based retrieval tasks: forward position$\rightarrow$item, backward position$\rightarrow$item, forward item$\rightarrow$position, and backward item$\rightarrow$position. LoRA reliably improves model performance over the base model across all four families. See Fig.~\ref{fig:fig5A} for per-query-position accuracy on the same data.}
  \label{fig:fig4}
\end{figure}

Having documented these failures in existing models, we now ask whether the deficits in position-based retrieval can be repaired through post-training. We constructed \textsc{PosBench}, a fine-tuning dataset generated by instantiating the four index operators (defined in Sec.~\ref{sec:results_setup}) over three sources of ordered structure: synthetic item sequences, code-line sequences from BigCode self-oss-instruct-sc2-exec-filter-50k~\citep{bigcode2023selfoss}, and naturally ordered fragments extracted from Open-Orca/SlimOrca~\citep{OpenOrca, SlimOrca}, OpenHermes-2.5~\citep{OpenHermes2.5}, and tiny-codes~\citep{tinycodes}. The resulting training mixture emphasized backward and relative addressing, the settings that expose the largest failures, while held-out evaluation samples were generated independently from the training examples. We then compared base models with LoRA and full-parameter supervised fine-tuning under answer-span supervision (the cross-entropy loss was computed only on tokens belonging to the answer span, with all other tokens masked out). Appendix~\ref{app:data-construction} details the construction of \textsc{PosBench}, while Appendix~\ref{app:data} lists the complete model, loss, and performance-evaluation definitions.

Fig.~\ref{fig:fig4} shows that LoRA fine-tuning improves position-based retrieval. Specifically, LoRA improves accuracy over the base model across all four model families. Evaluation was performed at $L=20$ on letter sequences with endpoint anchoring, averaged across the four position-based retrieval tasks (forward and backward position$\rightarrow$item and item$\rightarrow$position). Full-parameter fine-tuning does not yield similarly consistent gains across these families.

\subsection{Position-focused fine-tuning generalizes to held-out code understanding}\label{sec:results_pyindex}

To test whether the gains from position-focused fine-tuning generalize to code-understanding queries, we created \textsc{PyIndex}, a held-out benchmark in which each prompt is a Python list expression and the gold answer is the value the expression produces. The benchmark includes 5 subcategories: forward indexing (\texttt{xs[k]}), backward indexing (\texttt{xs[-k]}), nested indexing, where a list element is used as the index into the same list (\texttt{xs[xs[k]]}), expression-level indexing where the index is an arithmetic expression (\texttt{xs[a+b]}), and chained indexing across multiple retrieval steps (\texttt{sorted(xs)[k]}). \textsc{PyIndex} thus tests the model's ability to evaluate Python's indexing semantics.

\begin{figure}[t]
  \centering
  \includegraphics[width=\linewidth]{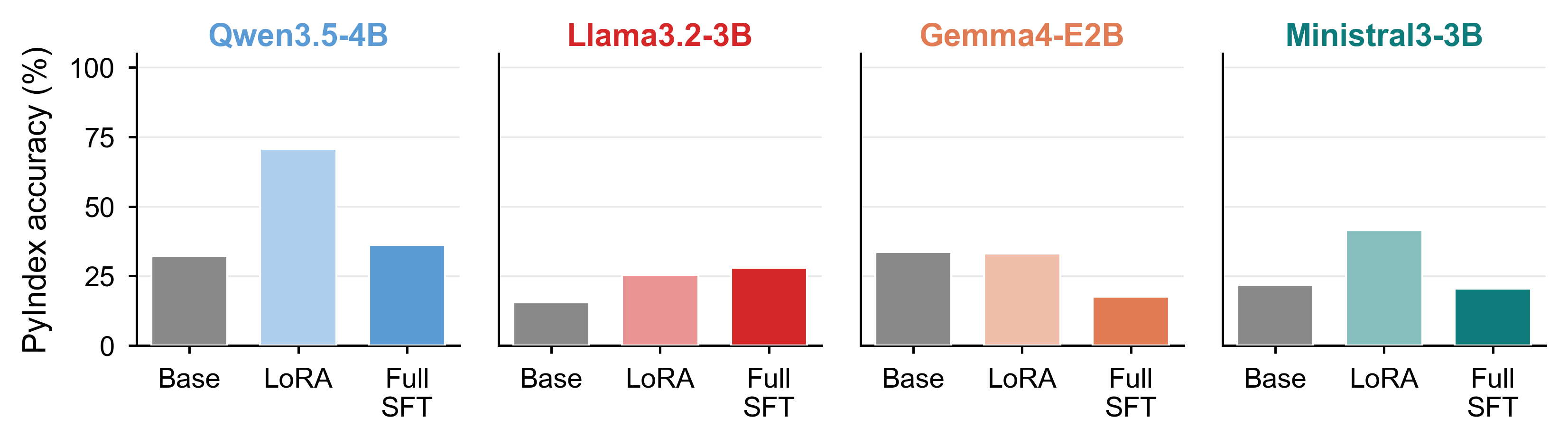}
  \caption{\textbf{Accuracy on the \textsc{PyIndex} dataset.}
  \textsc{PyIndex} accuracy across four open-source model families (Qwen3.5-4B, Llama 3.2-3B, Gemma4 E2B, Ministral 3.3B) under two fine-tuning methods (LoRA, full-parameter supervised fine-tuning SFT). Bars show mean accuracy over 100 held-out examples per model condition, equivalently the unweighted mean across \textsc{PyIndex}'s five subcategories with 20 examples each: Forward, Backward, Nested, Expression, and Chained. See Fig.~\ref{fig:figA4} for the per-subcategory breakdown.}
  \label{fig:fig5}
\end{figure}

Fig.~\ref{fig:fig5} reports average \textsc{PyIndex} accuracy across fine-tuning methods. On Qwen3.5-4B, the base model achieves $32.2\%$ accuracy. LoRA improves accuracy to $70.8\%$, more than doubling overall performance, whereas full-parameter fine-tuning improves only to $36.2\%$. This same pattern holds across the other backbones evaluated with LoRA. Qwen3.5-2B improves from $19.2\%$ to $31.4\%$, and Qwen3.6-35B-A3B improves from $43.8\%$ to $52.6\%$.

To understand which indexing operations benefited most, we analyzed performance by \textsc{PyIndex} subcategory in Appendix Fig.~\ref{fig:figA4}. The largest improvements appear in backward indexing (from $17\%$ to $71\%$), chained indexing (from $23\%$ to $72\%$), and expression-level indexing (from $22\%$ to $68\%$). Forward indexing improved to near-perfect accuracy (from $93\%$ to $99\%$). Finally, nested indexing also improves substantially (from $6\%$ to $44\%$), but remains the hardest subtask in absolute terms, suggesting that generalization is weakest when positional retrieval must interact with hierarchical structure. Together, these results show that the gains from position-focused fine-tuning extend beyond simple list indexing to richer Python expressions. The generalization is, however, only partial: \textsc{PyIndex} accuracy remains well below saturation, indicating that LoRA closes part of the position-based retrieval gap on code understanding but does not fully resolve it. This partial generalization likely reflects that the underlying position-based retrieval capability is not yet fully acquired during pretraining.
  
\section{Discussion}

This work introduces the \emph{Position Curse}: even as modern LLMs approach saturation on content-based retrieval over million-token contexts, they fail to retrieve items by position in short lists, with backward queries consistently worse than forward. The failure is not explained by counting, where models still report list length accurately on the same lists. Fine-tuning on \textsc{PosBench} closes part of the gap and generalizes to a held-out code-understanding benchmark (\textsc{PyIndex}), yet absolute performance remains far from saturated. This raises the question of \emph{why} position-based retrieval, especially its backward variant, is so much harder than its content-based counterpart.

We speculate on a possible mechanistic interpretation of why counting succeeds while backward position-based retrieval fails. In our experiments, models answer counting queries with high accuracy on the list lengths we consider, and they do so directly, without an explicit chain-of-thought enumeration. This indicates that the count of items is already present in the residual stream by the time the model emits its answer, rather than being derived through step-by-step decoding. This pre-computed count is consistent with known mechanistic accounts. As the model autoregressively reads each new item, a hidden state tracker can write the running ordinal index into that item's residual stream, effectively tagging each token with ``I am the $i$-th item.'' Two recent lines of work make this picture concrete. \citet{gurnee2026geometryofcounting} shows that Claude 3.5 Haiku represents character counts on a low-dimensional curved manifold that is incrementally updated as the model reads new tokens. \citet{feng2024bindingentities} identifies a binding-ID mechanism by which language models attach a position-like identifier vector to each in-context entity, so that an item and its forward position become bound in the same activation. To answer a counting query, the model can therefore locate the end of the list and read off the position information already carried by that final item. The same forward-position information can also support forward position-based retrieval. For an item$\rightarrow$position query, the queried item in the prompt can be matched to its earlier occurrence in the list by a duplicate-token head~\citep{wang2023interpretabilityioi}, and the bound position vector can then be read out of the residual stream of that earlier occurrence.

Backward retrieval, in contrast, requires a position subtraction rather than a direct readout. A fully general algorithm would have to first locate the target item's position, then locate the list-end item's position, and finally compute the difference between the two. Under rotary position embeddings (RoPE)~\citep{su2024roformer}, however, position information enters only through the rotation of attention keys and queries; it is never written into the value projection and therefore never appears in the residual stream itself. This makes the subtraction of two positions difficult to implement with a generic circuit. As a consequence, the model likely relies on less generalizable shortcuts acquired during pretraining when the query requires backward indexing, which would explain the consistent forward--backward asymmetry we observe and the partial recovery achieved by full-parameter fine-tuning. Together, these observations suggest that robust position-based retrieval and access may require either new pretraining data or new model architectures explicitly designed to support such operations.

\paragraph{Limitations and future work.} A limitation of this work is that our evaluation focuses on a targeted positional benchmark  (\textsc{PosBench}, \textsc{PyIndex}); extending the analysis to broader settings is a natural next step. Designing post-training procedures that yield generalizable position-based retrieval, rather than only in-distribution gains, is an important direction for future work.

\paragraph{Acknowledgements.} We thank collaborators and colleagues for discussions and feedback. Z.Z was supported in part by the Halıcıoğlu Data Science Institute (HDSI) Ph.D. Fellowship. Z.Z especially thanks Gal Mishne for her continued mentorship and support. 
The authors gratefully acknowledge support from the Tinker Research Grant and the Center of Excellence in Computational Cognition Pilot Grant, both awarded to H.-D.~X.
 
\bibliographystyle{plainnat}
\bibliography{ref}

\newpage
\appendix
\raggedbottom



\section{Prompts for Figure~\ref{fig:fig1}}
\label{app:fig1-prompts}

The two failure cases shown in Fig.~\ref{fig:fig1}(b) and Fig.~\ref{fig:fig1}(c) use the prompts below, sent verbatim to Claude Opus 4.6. Both cases were queried through the web interface and the API; for the API case, we used temperature $0$ with reasoning disabled.  

\paragraph{Letter retrieval (Fig.~\ref{fig:fig1}b).}
\begin{verbatim}
Below is a sequence of letters. What is the second-to-last letter?
Respond with ONLY that single letter, nothing else.

X, V, Z, Y.
\end{verbatim}

\paragraph{Code branch retrieval (Fig.~\ref{fig:fig1}c).}
\begin{verbatim}
Below is a Python if-elif branch. Your task is to repeat the
second-to-last branch exactly as written (without the leading
whitespace). Respond with ONLY that single branch, nothing else.

if status == 200:
    return "OK"
elif status == 404:
    return "Not Found"
\end{verbatim}

\section{Other letter and word results}
\begin{figure}[H]
    \centering
    \includegraphics[width= \linewidth]{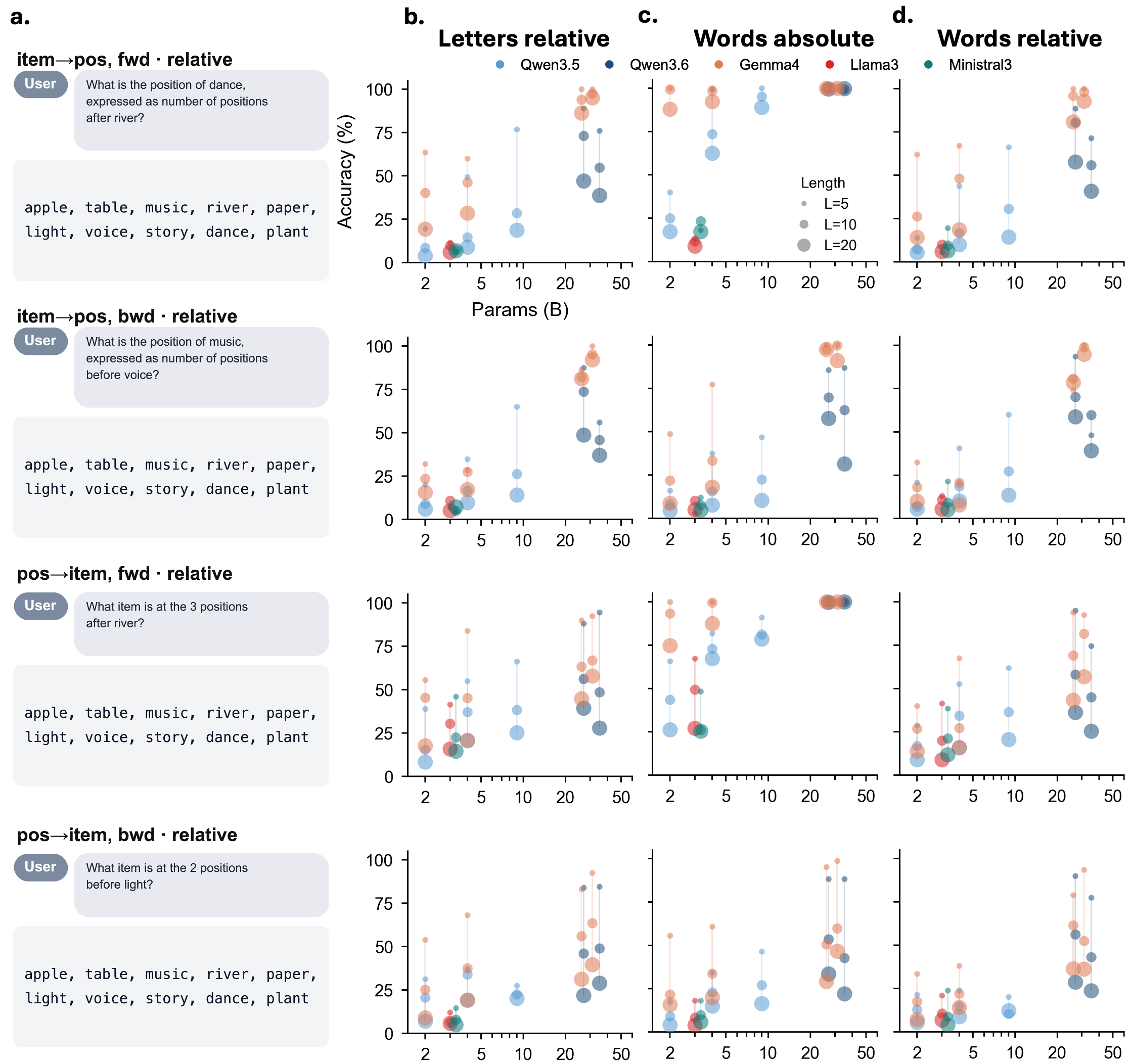}
    \caption{\textbf{Position-based retrieval across tasks and models.}
    \textbf{(a)} Example prompts for the task conditions plotted in panels (b)--(d).
    \textbf{(b)} \emph{Letters Relative}: the list contains letters, and the queried position is relative to another letter in the list (e.g., ``the letter that comes two positions after V'').
    \textbf{(c)} \emph{Words Endpoint}: the list contains words, and the queried position is relative to the start or end of the list (e.g., ``the second-to-last word'').
    \textbf{(d)} \emph{Words Relative}: the list contains words, and the queried position is relative to another word in the list (e.g., ``the word that comes two positions after \texttt{apple}''). Panels (b)--(d) report mean exact-match accuracy across model scales and sequence lengths, averaged over valid query positions and sampled trials within each task condition. Color denotes model family, and marker sizes from smallest to largest correspond to sequence lengths $L=5$, $10$, and $20$.}
        
    \label{fig:figA2}
\end{figure}
\section{Methods and Evaluation Details}
\label{app:data}

\subsection{Models}
\label{app:models}

The retrieval sweeps cover the model families reported in Fig.~\ref{fig:fig2}: Qwen3.5 dense models at 2B, 4B, and 9B parameters; Qwen3.6-35B-A3B; Gemma-family models; and proprietary Claude and GPT systems where available. Unless a figure specifies otherwise, fine-tuning experiments instantiated Qwen3.5-4B as the backbone, enabling us to compare LoRA and full-parameter supervised fine-tuning (SFT) under matched data and training conditions. The \textsc{PyIndex} generalization analysis extends this comparison to Qwen3.5-4B, Llama 3.2-3B, Gemma4 E2B, and Ministral 3.3B. Reasoning is disabled by default and enabled only in the explicitly labeled reasoning comparison.



\subsection{Position-focused data construction}
\label{app:data-construction}

The SFT corpus combines three sources of ordered structure,
\[
\mathcal{D}_{\mathrm{SFT}}=\mathcal{D}_{\mathrm{syn}}\cup\mathcal{D}_{\mathrm{code}}\cup\mathcal{D}_{\mathrm{adapt}},
\]
with 20K synthetic retrieval examples, 4K code-retrieval examples, and 46K adapted real-data examples. Across generated retrieval examples, we sample $P(\mathrm{forward})=0.3$ and $P(\mathrm{endpoint})=0.3$, intentionally overrepresenting backward and relative addressing because these conditions most strongly expose the Position Curse.

The synthetic component provides controlled coverage of the operator family. It samples both Position$\rightarrow$Item and Item$\rightarrow$Position queries over endpoint and relative anchors, with sequence lengths drawn from $[10,50]$. Items come from categorical pools such as letters, digits, animals, fruits, cities, elements, languages, and instruments. We randomize list formatting and query phrasing, and include a small number of multi-turn list conversations so that the model sees positional queries in more natural dialogue contexts.

The code component uses self-oss-instruct-sc2-exec-filter-50k~\cite{bigcode2023selfoss} to expose the same operator to structured code spans. Each usable snippet is split into non-empty lines; snippets with fewer than five lines are discarded, and longer snippets are windowed to contiguous spans of 5--30 lines before positional queries are instantiated over the resulting line sequence.

The adapted component increases linguistic diversity by converting naturally ordered structures from Open-Orca/SlimOrca~\cite{OpenOrca,SlimOrca}, OpenHermes-2.5~\cite{OpenHermes2.5}, and tiny-codes~\cite{tinycodes} into retrieval examples. We extract numbered lists, bullet lists, markdown tables, and code blocks when they contain at least five usable items, rows, or lines. Extracted items are deduplicated, constrained to short spans, and reformatted before query generation. For tiny-codes, we preserve the original dialogue context and append the positional retrieval query as a follow-up turn.
   
\subsection{Evaluation splits and tasks}
\label{app:evaluation}
 
\textsc{PosBench} evaluation uses held-out sequences sampled independently from the training corpus. Each condition fixes the query type, anchor type, indexing direction, item type, prompt variant, and sequence length $L\in\{5,10,20\}$. We sample 50 independent sequences per condition and instantiate the valid offsets for each sequence, yielding deterministic targets while preserving variation in the list content. Every test prompt is preceded by three in-context demonstrations of the same operator drawn from independently sampled sequences, so all reported numbers correspond to three-shot evaluation; this disambiguates the task format (e.g., indexing convention and answer phrasing) so that performance reflects position-based retrieval rather than task-comprehension noise. Unless otherwise specified, generations use temperature $0.7$ and are scored by exact match after task-specific parsing.


\begin{figure}[htbp]
    \centering
    \includegraphics[width=\linewidth]{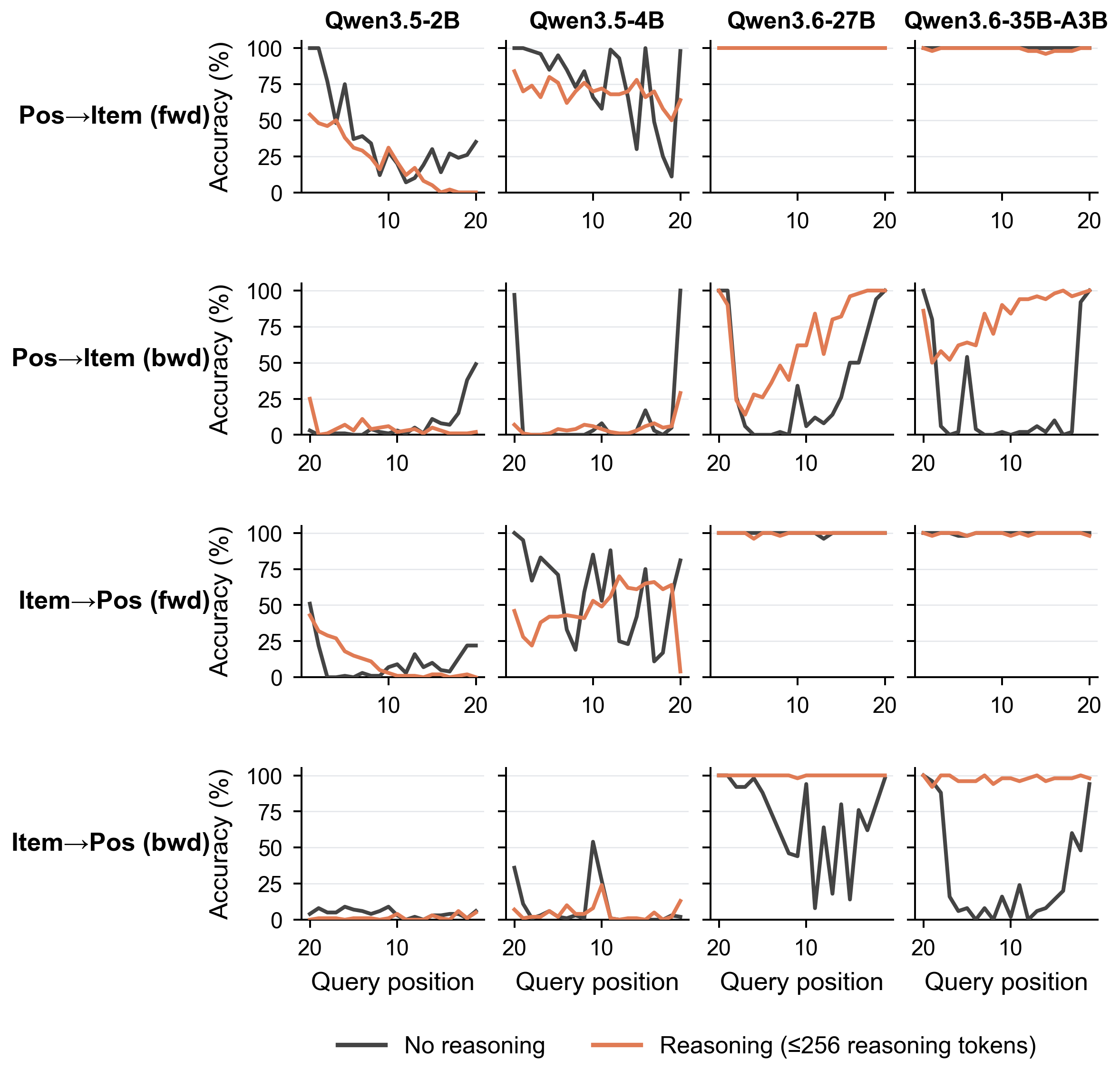}
    \caption{\textbf{Reasoning does not consistently rescue position retrieval.}
    Per-query-position accuracy at sequence length $L{=}20$ on the four position-retrieval tasks, comparing each Qwen model under two conditions: no-reasoning (black; the model produces the answer directly, without emitting any intermediate reasoning tokens) and reasoning (orange; the model is allowed to emit up to 256 reasoning tokens before producing the final answer). Each plotted value is the mean exact-match accuracy across sampled trials at that queried position. Rows correspond to the four task variants, and columns correspond to the four models tested. }
    \label{fig:figA1}
\end{figure}

\subsection{Answer-span supervision and fine-tuning methods}
\label{app:adaptation}

Retrieval examples apply supervision only to the answer span rather than to the full assistant message. Let $m_t\in\{0,1\}$ indicate whether token $t$ belongs to the target answer. The retrieval loss is
\[
\mathcal{L}_{\mathrm{retrieval}}=-\sum_t m_t \log p_\theta(y_t\mid y_{<t},x).
\]
This masking prevents prompt text and optional answer framing from becoming supervised targets. 

All fine-tuning experiments use bf16 precision, maximum sequence length 512, cosine learning-rate decay with 3\% warmup, and seed 42. The LoRA method applies rank-$32$ adapters with $\alpha=32$ and dropout $0$ to the attention projections $(q,k,v,o)$ and the MLP projections (gate, up, down); it trains for one epoch with learning rate $10^{-4}$, weight decay $0.05$, per-device batch size 16, and gradient accumulation 2. Full-parameter fine-tuning updates all parameters for 0.2 epochs with learning rate $2\times10^{-5}$, weight decay $0.1$, per-device batch size 4, gradient accumulation 8, gradient clipping at 1, and NEFTune noise $\alpha=5$. 

\subsection{Reasoning}
For each Qwen model that supports reasoning (Qwen3.5-2B, 3.5-4B, Qwen3.6-27B, Qwen3.6-35B-A3B), we evaluated the same position-retrieval benchmark twice using the local Hugging Face backend while keeping the model weights, prompts, sampled sequences, and decoding seed fixed. In the \emph{no-reasoning} condition, the model generated an answer directly from a single forward pass over the prompt. In the \emph{reasoning} condition, we enabled the model's built-in reasoning channel (\texttt{enable\_thinking=True}) with a budget of 256 reasoning tokens; the model first generated a reasoning trace and then produced a final answer. Sequences, prompt phrasing, endpoint anchor, and all four task variants (Pos$\rightarrow$Item forward/backward, Item$\rightarrow$Pos forward/backward) were identical across the two conditions, so any difference in per-position accuracy is attributable to the reasoning trace rather than changes in prompts or samples. Accuracy was computed independently for each query position using 100 trials per (task, position) cell in Fig.~\ref{fig:figA1}.

\subsection{\textsc{PyIndex} benchmark}
\label{app:pyindex}
   
\textsc{PyIndex} isolates code indexing with short Python snippets containing a list literal and a single indexing expression. Let $X=(x_0,\ldots,x_{L-1})$ denote the list. The benchmark spans five subcategories. Forward and Backward evaluate direct positional retrieval under forward indexing ($X[i]$) and reverse indexing ($X[-i]$), respectively. Nested evaluates multi-step positional composition (e.g., $X[X[i]]$), where retrieval must be performed relative to an intermediate retrieved item. Expression evaluates arithmetic-style positional expressions (e.g., $X[a+b]$, $X[L-k]$, $X[a\bmod L]$, or $X.\mathrm{index}(v)$). Chained evaluates sequential retrieval operations composed across multiple indexing steps (e.g., $X[a:b][i]$, $\operatorname{reversed}(X)[i]$, or $\operatorname{sorted}(X)[i]$).

Gold answers are computed directly from the generated Python expression, so each example tests whether the model can resolve the positional transformation rather than synthesize a program. Each model is evaluated on 20 examples per category using a fixed seed, yielding 100 total held-out trials per model. We report both the unweighted mean across categories (Fig.~\ref{fig:fig5}) and the per-category breakdown (Fig.~\ref{fig:figA4}, shown below).

\begin{figure}[t]
  \centering
  \includegraphics[width=\linewidth]{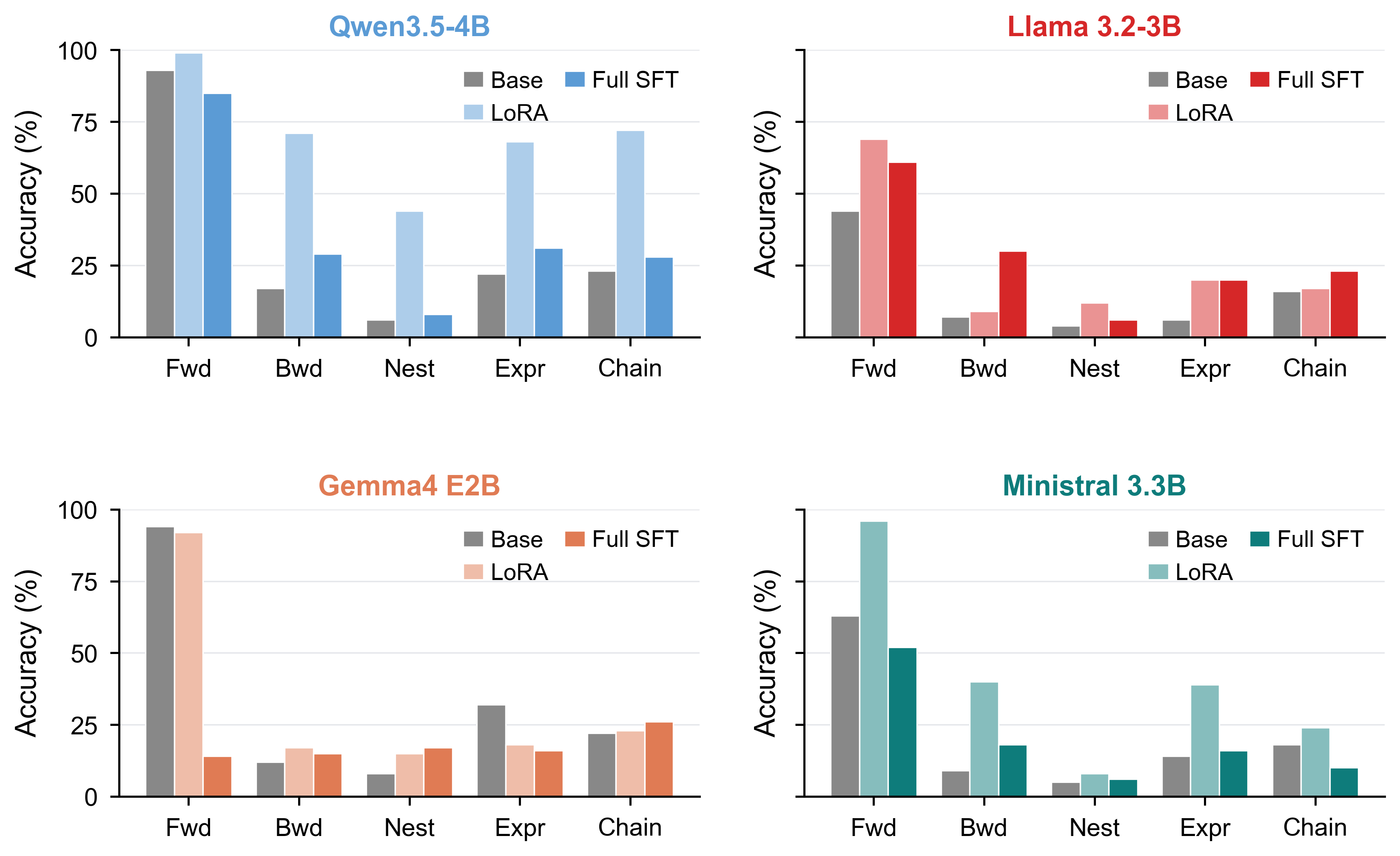}
  \caption{\textbf{\textsc{PyIndex} accuracy by subcategory (detailed view of Fig.~\ref{fig:fig5})}
  Per-subcategory breakdown of the same data summarized in Fig.~\ref{fig:fig5}: \textsc{PyIndex} accuracy across four open-source model families (Qwen3.5-4B, Llama 3.2-3B, Gemma4 E2B, Ministral 3.3B) under two SFT methods (LoRA, Full-Parameter Fine-Tuning), shown separately for \textsc{PyIndex}'s five subcategories: Forward, Backward, Nested, Expression, and Chained. Each bar averages 20 held-out examples from one subcategory and model condition.}
  \label{fig:figA4}
\end{figure}

\subsection{Performance evaluation}
\label{app:scoring}

For each trial $t$, the generated response is first mapped into the appropriate answer space and then compared with the deterministic target $y^{(t)}$. Item-retrieval responses are matched against the candidate items in the sequence. Position and count responses use the first integer answer, and \textsc{PyIndex} responses are normalized by removing formatting artifacts such as code fences, quotes, and backticks before short-answer matching. For any subset of trials $\mathcal{T}$, accuracy is
\[
\operatorname{Acc}(\mathcal{T})=\frac{1}{|\mathcal{T}|}\sum_{t\in\mathcal{T}}\mathbf{1}[\hat{y}^{(t)}=y^{(t)}].
\]
Per-offset summaries use the subset $\mathcal{T}_n=\{t:n^{(t)}=n\}$.

\subsection{Compute resources}
\label{app:compute}

Local open-source experiments used H100 80GB GPUs. All local runs except Qwen3.6-35B-A3B used one H100 80GB, while Qwen3.6-35B-A3B used two H100 80GB GPUs. Standard inference took roughly 5--8 hours per model, reasoning-mode inference roughly 20 hours per model, and SFT roughly 0.5--2 hours per method. For the current result set (11 non-reasoning local inference sweeps, 8 reasoning sweeps, and 15 SFT runs), this corresponds to approximately 250--310 single-H100-equivalent hours. API-backed evaluations or managed runs used the OpenAI API, Claude API, and Tinker API rather than local GPU accounting.

\section{Position Retrieval Tasks with SFT}
\begin{figure}[htbp]
    \centering
    \includegraphics[width=\linewidth]{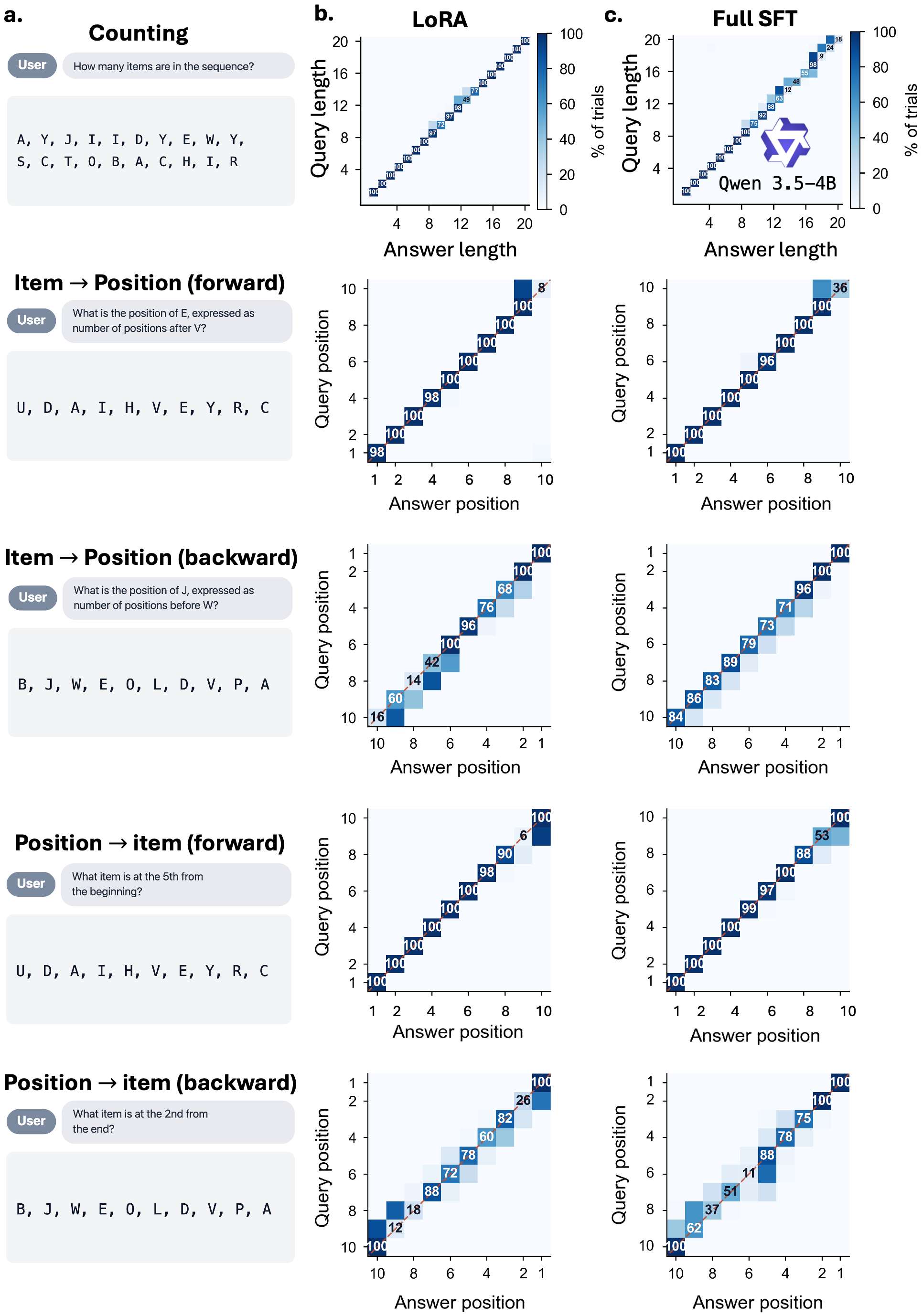}
    \caption{\textbf{Position-based Retrieval with Lora and SFT.}
    \textbf{(a)} Same example prompt and sequence for each task as Fig \ref{fig:fig2}a, shown for clarity.
    \textbf{(b)} Per-task accuracy of Qwen3.5-4B fine-tuned with LoRA, shown as column-normalized confusion heatmaps (cell color = \% of trials within each query bin; red dashed diagonal indicates correct answers). For counting, axes denote query and answer \emph{length}. For retrieval tasks, axes denote query and answer \emph{position}. Diagonal cells thus reflect per-length or per-position accuracy.
    \textbf{(c)} Per-task accuracy of Qwen3.5-4B after full-parameter supervised fine-tuning SFT.}
    
    \label{fig:fig3A}
\end{figure}

We showed the examples of task accuracy with LoRA (Fig.~\ref{fig:fig3A}b) and full-parameter fine-tuning (Fig.~\ref{fig:fig3A}c) in corresponding tasks. 

\begin{figure}[htbp]
    \centering
    \includegraphics[width=\linewidth]{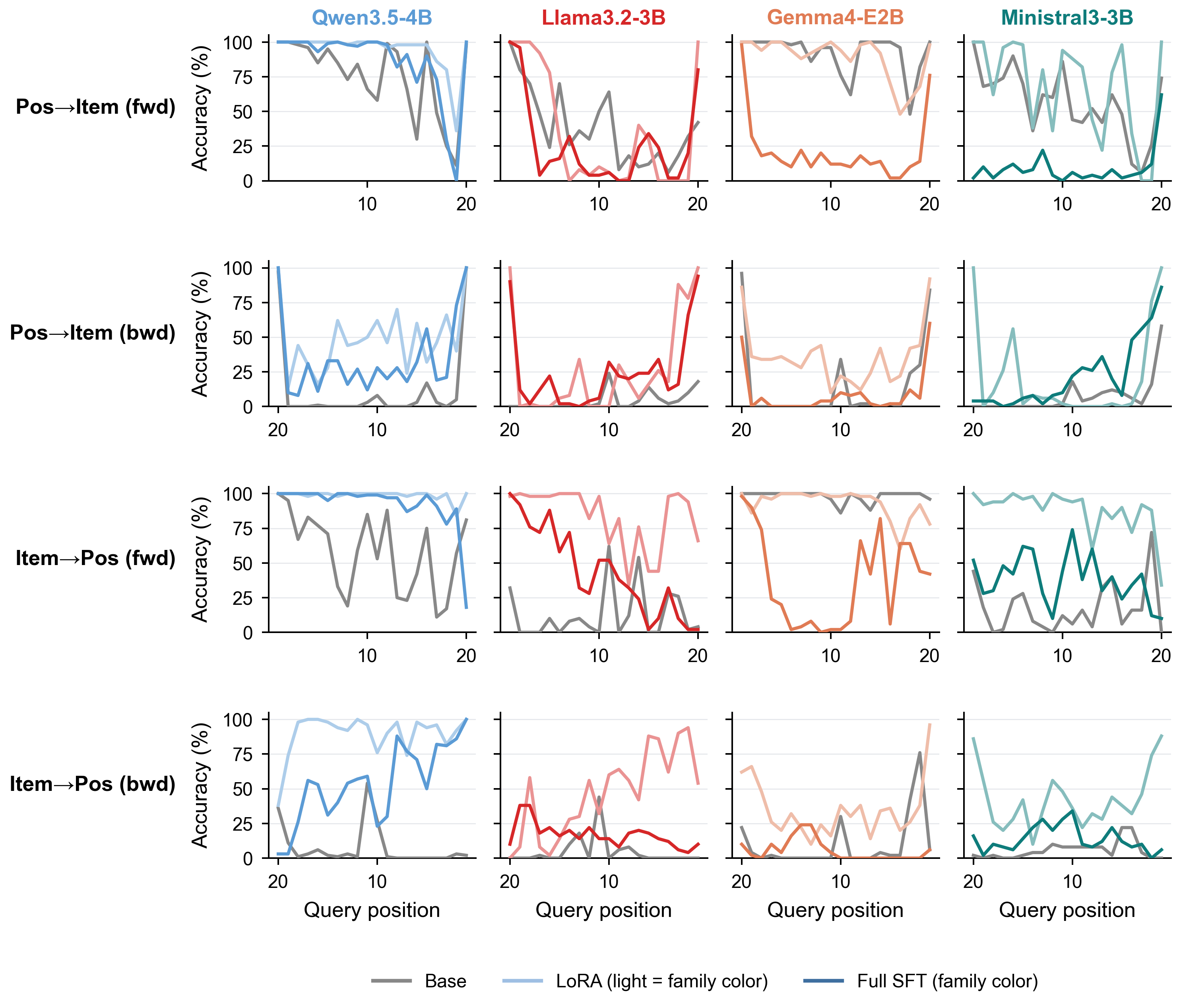}
    \caption{\textbf{Per-position accuracy by fine-tuning method.}
    Per-query-position accuracy at $L{=}20$ for the four position-retrieval tasks and four open-source models in Fig~\ref{fig:fig4}. Each plotted value is the mean exact-match accuracy across sampled trials at that queried position. Rows denote task variants; columns denote model families. In each panel, gray shows the pre-trained base model, light color shows LoRA, and dark color shows full-parameter fine-tuning. Backward-query panels are ordered by absolute position (decreasing left-to-right), matching the directional convention used in Fig.~\ref{fig:fig2}b,d. All results use letter sequences, and endpoint anchoring.}

    \label{fig:fig5A}
\end{figure}

\end{document}